\documentclass{article} 
\usepackage{iclr2023_conference_tinypaper,times}

\usepackage{amsmath,amsfonts,bm}

\def\eqref#1{equation~\ref{#1}}

\def\1{\bm{1}}

\DeclareMathAlphabet{\mathsfit}{\encodingdefault}{\sfdefault}{m}{sl}
\SetMathAlphabet{\mathsfit}{bold}{\encodingdefault}{\sfdefault}{bx}{n}

\usepackage{hyperref}
\usepackage{url}
\usepackage{svg}
\usepackage{graphicx}
\usepackage{multirow}
\usepackage{adjustbox}
\usepackage{tabularx}

\usepackage{subcaption}

\title{VoltaVision: A Transfer Learning model for electronic component classification}

\author{Anas Mohammad Ishfaqul Muktadir Osmani, Taimur Rahman \& Salekul Islam\\
Department of Computer Science and Engineering\\ 
United International University\\
{\small\texttt{\{aosmani203004,trahman221427\}@bscse.uiu.ac.bd,salekul@cse.uiu.ac.bd}} \\
}

\iclrfinalcopy 
\begin{document}

\maketitle

\begin{abstract}
 
In this paper, we analyze the effectiveness of transfer learning on classifying electronic components. Transfer learning reuses pre-trained models to save time and resources in building a robust classifier rather than learning from scratch. Our work introduces a lightweight CNN, coined as VoltaVision, and compares its performance against more complex models. We test the hypothesis that transferring knowledge from a similar task to our target domain yields better results than state-of-the-art models trained on general datasets. Our dataset and code for this work are available at \url{https://github.com/AnasIshfaque/VoltaVision}.
\end{abstract}

\section{Introduction}



Traditional transfer learning uses large pre-trained models on general classification tasks to cut down on the time required for training. However, it may be more valuable to use purpose-built transfer models on focused tasks. Focused tasks such as medical imaging, waste classification and facial recognition have become far more relevant over a one-size-fits-all solution. One such problem is detecting electronic components. Most electronic components are minute and intricate in design which means a rich diverse dataset would be needed to capture their defining features. In order to mitigate the effort of creating such a dataset, transfer learning can be used to train a reliable image classifier needing only a few samples of the target class. 

The applications of identifying small electronic components would be significant in recycling e-waste, or automating component transactions in a lending machine as prototyped in Appendix \ref{sec: hardware setup}. The benefit of this approach would be to reduce model footprint considerably while maintaining accurate inference.

\section{Related Works}
Siamese networks, also known as twin neural networks, have been used for few-shot learning to identify electronic components \citep{siamesenetwork}. They use an improved VGG-16 \citep{simonyan2014very} model and a visual geometry group with five convolutional layers for feature extraction to improve recognition even with smaller samples. In \citep{deeptronic}, it is shown that deep convolutional neural networks can be used with transfer learning to identify electronic components. Pre-trained models such as Inception-v3 \citep{szegedy2016rethinking}, GoogLeNet \citep{szegedy2015going} and ResNet-101 \citep{he2016deep} are used for faster deployment. A faster SqueezeNet network is used in \citet{xu2020n} as the basis for developing a simpler classification algorithm having fewer parameters without compromising the performance.

\section{Methodology}
Our target task is to classify three electronic components; humidity sensor (DH11), transistor and Bluetooth module. To achieve this with transfer learning, we have fine-tuned various pre-trained models using a small dataset of our target components, described in Appendix \ref{sec: dataset}. We have created this dataset adopting the methodology used in \citep{chand2023low} by capturing images of the components from various view angles. The pre-trained models are either trained on a general dataset or a specific dataset that belongs to the same domain as the fine-tune data.

We have experimented with some large well-known image classifiers as the pre-trained model which includes the ResNet-18 \citep{He_2016_CVPR}, VGG-16 \citep{simonyan2014very}, Inception-v3 \citep{szegedy2016rethinking}, AlexNet \citep{krizhevsky2012imagenet} and GoogLeNet \citep{szegedy2015going}. These models contain weights trained on ImageNet \citep{ILSVRC15}. We train our model, VoltaVision with general CIFAR datasets \citep{krizhevsky2009learning}, and also task-relevant datasets \citep{chand2023low, elecomp} with varying class sizes to be used as the pre-trained model (shown in Appendix \ref{sec: custom CNN}). Figure \ref{method_diagram} outlines our experimental setup, and more technical details are covered in Appendix \ref{sec: experiment}. Since our fine-tuning dataset is relatively small, we perform 5-fold cross validation to ensure that our reported performance values are accurate. 

\begin{center}
    \begin{figure}[ht!]
    \centering
    \begin{adjustbox}{max width=\textwidth}
        \includegraphics{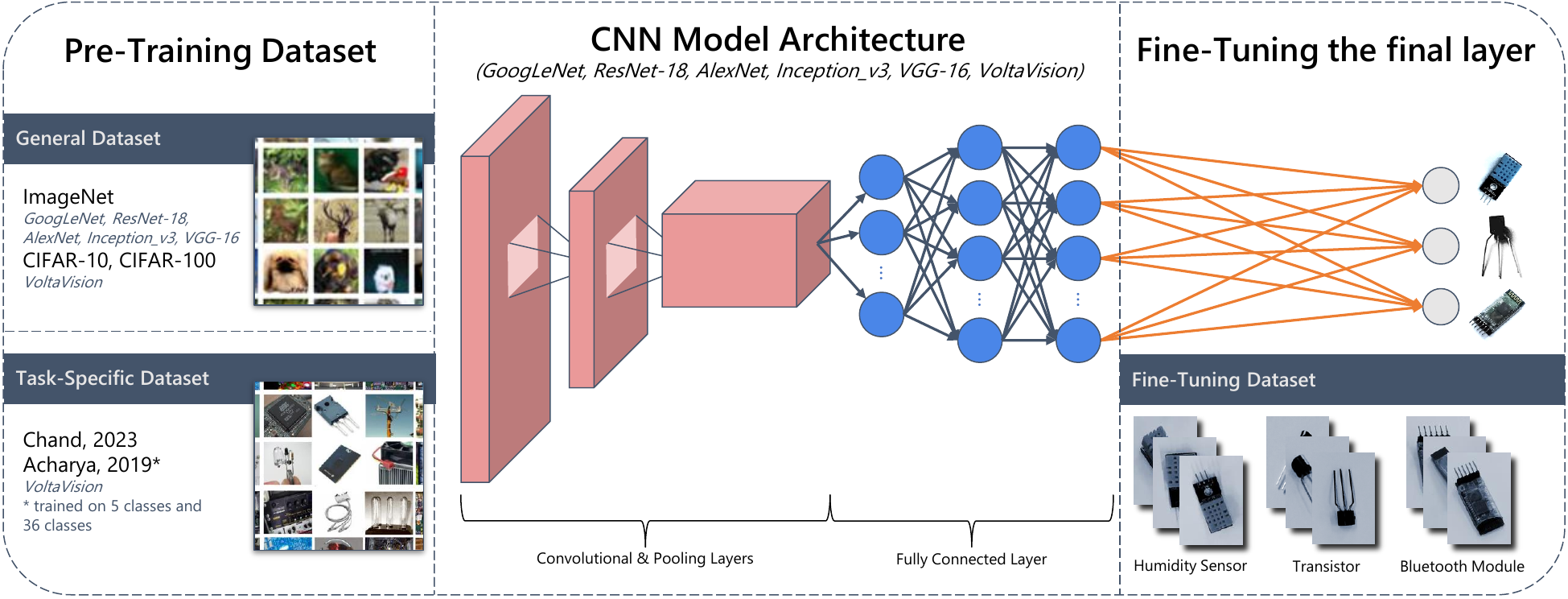}
    \end{adjustbox}
    \caption{Experimental setups\label{method_diagram}}
    \end{figure}
\end{center}


\section{Results and Discussion}

\begin{table}[!htbp]
\begin{center}

\begin{adjustbox}{width=\textwidth}
\begin{tabular}{|l|c|c|c|c|c|c|c|c|}
\hline 
\textbf{Pre-Train Dataset} & \textbf{Model}   & \textbf{Accuracy}     & \textbf{Precision}    & \textbf{Recall}    & \textbf{F1-Score}   & \textbf{Fine-tuning Time} & \textbf{GPU Usage} &\textbf{Model Size}\\ \hline
\multirow{5}{*}{ImageNet\textsubscript{1000} }&GoogLeNet &78.00    &78.65     &78.29  &77.94   &84.1&225.1& 49.7Mb\\ \cline{2-9} 
&ResNet-18               &80.60    &81.20     &80.76  &80.67 &77.9&181.5&44.7Mb   \\ \cline{2-9}
&Inception-v3            &82.60    &83.31     &82.90  &82.78 &121.1&521.2&104Mb   \\ \cline{2-9}
&AlexNet                 &96.40    &96.38     &96.35  &96.32 &66.9&844.8&233Mb   \\ \cline{2-9}
&VGG-16                  &\textbf{98.80}    &\textbf{98.78}     &\textbf{98.85}  &\textbf{98.81}  &117.5&1968&528Mb  \\ \hline

CIFAR-10\textsubscript{10} &\multirow{5}{*}{VoltaVision} &76.40&	76.80&	76.02&	75.51 &53.1&31.1&320kb\\ \cline{1-1}\cline{3-9}

CIFAR-100\textsubscript{100} & &89.60&	89.85&	89.67&	89.73 &48.4&23.6&2.92Mb\\ \cline{1-1}\cline{3-9}
\citet{chand2023low}\textsubscript{3} &  &95.20    &95.21     &95.25  &95.20 &46.7&\textbf{18}&\textbf{127kb}\\ \cline{1-1}\cline{3-9}
\citet{elecomp}\textsubscript{5} &       &93.60    &93.85     &93.61  &93.67   &\textbf{45.5}&18.1&185kb  \\ \cline{1-1}\cline{3-9}
\citet{elecomp}\textsubscript{36}&       &92.60    &92.31     &92.64  &92.38   &46.8&18.9&1.08Mb  \\ \hline
\end{tabular}
\end{adjustbox}
\caption{Average performance metrics for corresponding pre-trained models with different datasets, where subscripts denote the number of classes. Performance metrics are shown in percentages, time is in seconds, GPU usage is in MiB.}
\label{result_summary}
\end{center}
\end{table}

Our average results are summarized in Table \ref{result_summary}. It is noticeable that our model, VoltaVision (a custom CNN model) shows performance comparable to that of VGG-16 and AlexNet when pre-trained on a task-oriented dataset similar to our target. VoltaVision achieves this while it is smaller in size, requires less computational resources, and is faster in training. Furthermore, it is worth noting that when VoltaVision is pre-trained using general datasets CIFAR-10 and CIFAR-100 the results from a task-specific pre-trained model is still relatively higher. Thus, it shows that a more focused knowledge transfer yields better results in most cases (i.e. excluding complex architectures). 



For future work, there is a scope to classify more components, or consider further datasets for pre-training, as well as experiment with other model architectures. It may also be worthwhile to consider finer classifications, such as subcategories within transistors. Finally, it might be interesting to see if fine-tuning layers beyond just the final one yield different results. 




\section*{URM Statement}
The authors acknowledge that at least one key author of this work meets the URM criteria of ICLR 2024 Tiny Papers Track.

\bibliography{iclr2023_conference_tinypaper}
\bibliographystyle{iclr2023_conference_tinypaper}

\appendix
\section{Appendix}

\subsection{Electronic Lending Machine}

\label{sec: hardware setup}
\begin{center}
    \begin{figure}[!htbp]
    \centering
    \begin{adjustbox}{max width=\textwidth}
        \includegraphics{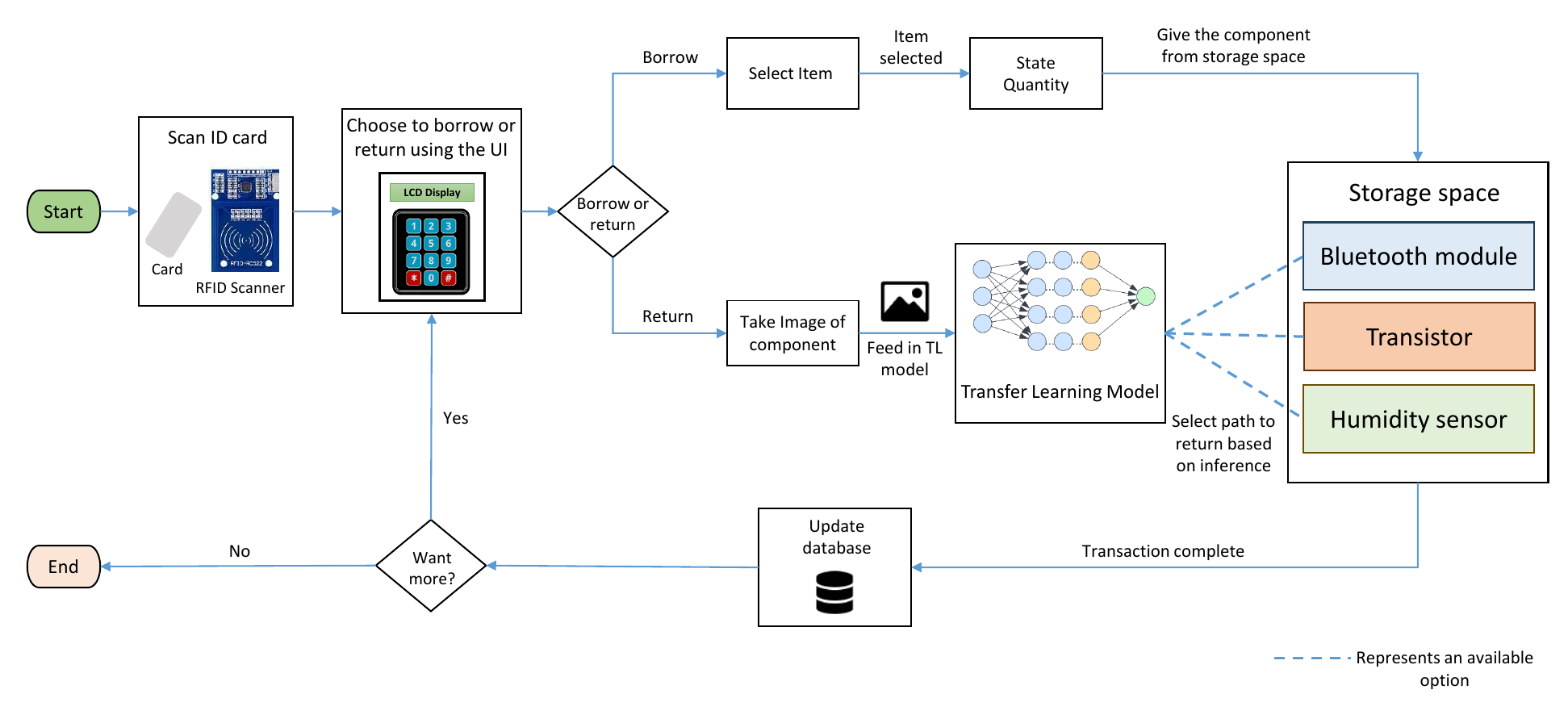}
    \end{adjustbox}
    \caption{Overview of the electronic lending machine prototype  \label{machine_overview}}
    \end{figure}
\end{center}

Figure \ref{machine_overview} shows an overview of the hardware implementation of the prototype we build. There are two major functions of the machine; borrowing and returning components. An authentication mechanism is deployed using an RFID module, where the user needs to scan a registered card on the RFID scanner to get access to the system. For interacting with the machine, a UI module is designed consisting of a 16x2 LCD display and a 3x4 keypad. For automating the component transaction process, several servo motors and stepper motors are used for precise handling of the components. The authentication module, UI module and the actuators (i.e., the motors) are interfaced with an Arduino MEGA microcontroller which is also connected to a Windows laptop. For tracking the lending records, a database server is maintained, which is updated after every transaction.

The transfer learning model is used when returning a component. An image of the component to be returned is captured via a webcam and is fed into the model for inference. In response to this prediction, a sequence of servo motor rotation moves the component from the photo-shooting platform to its designated slot in the storage space. 

\begin{center}
    \begin{figure}[!htbp]
    \centering
    \includegraphics[width=100mm]{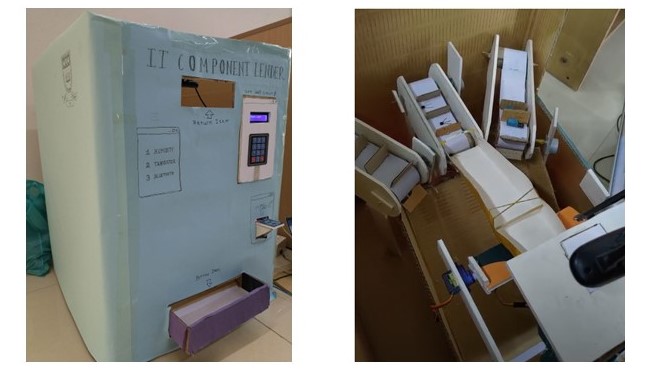}
    \caption{Lending machine from user view and internal view\label{machine}}
    \end{figure}
\end{center}

\subsection{Dataset Description}
\label{sec: dataset}

\begin{table}[h]
	\begin{subtable}[h]{0.48\textwidth}
		\centering
		\begin{tabular}{|c|c|}
                \hline
                \textbf{Classes} & \textbf{Instances}\\ \hline
                Transistor & 110\\  \hline
                Humidity Sensor &116\\ \hline
                Bluetooth Module & 102\\ \hline
            \end{tabular}
            \caption{Our dataset description}
            \label{tab:label subtable A}
	\end{subtable}
    \begin{subtable}[h]{0.48\textwidth}
		\centering
		\begin{tabular}{|c|c|c|}
                \hline
                \textbf{Dataset} & \textbf{Classes}&\textbf{Total Images}\\ \hline
                ImageNet-1K& 1,000& 1,281,167\\  \hline
                CIFAR-100& 100 &60,000\\ \hline
                CIFAR-10& 10 & 60,000 \\ \hline
                \citet{chand2023low}& 3& 1,810\\ \hline
                \citet{elecomp} & 36& 10,990 \\ \hline 
            \end{tabular}
            \caption{Datasets used for pre-training}
            \label{tab:label subtable B}
	\end{subtable}
	\caption{Description of datasets used in our experiments}
	\label{dataset description}
\end{table}

For building our fine-tuning dataset, we take pictures of three distinct electronic components from different angles on a white background. There are over 100 images for each of the classes, totalling to 328 images. The dataset is available at our GitHub\footnote{\url{https://github.com/AnasIshfaque/VoltaVision}} and it is described in Table \ref{tab:label subtable A}. Furthermore, readily available pre-trained models such as AlexNet are trained on ImageNet \citep{ILSVRC15}. While it is not feasible for us to train VoltaVision on a large dataset such as ImageNet, VoltaVision is pre-configured with weights trained from CIFAR \citep{krizhevsky2009learning}, \citet{chand2023low} and \citet{elecomp}. These pre-training datasets are outlined in Table \ref{tab:label subtable B}. Another experiment is carried out with a manually downsized \citet{elecomp} dataset to five classes (electrolytic capacitor, electric relay, heat sink, potentiometer and solenoid).

\subsection{Building VoltaVision by customising CNN model}
\label{sec: custom CNN}
The term “Volta” is derived from voltage as it is closely related to every electronic component and “Vision” comes from the fact that our model works with visual data. In terms of the network architecture, there are 3 convolution layers. The first convolution layer gets the image as the input and has 12 filters of size 3, stride 1 and padding 2. Then, batch normalization and ReLU activation are applied. To reduce the dimension of the feature map, a max pooling layer is used next. The second and third convolution layers have 20 and 32 filters of the same size as the first one. Similarly, the outputs of these convolution layers have also gone through batch normalization and ReLU activation. Finally, the output of the last convolutional layer is flattened and fed to a fully connected layer as a vector. 

\subsection{Transfer Learning Experiments}
\label{sec: experiment}
\begin{table}[!htb]
    \centering
    \begin{tabular}{|c|c|c|}
    \hline
    \textbf{Model} & \textbf{Parameters} & \textbf{Number of Layers} \\ \hline
    VGG-16 & 138 million & 16 \\ \hline
    Inception-v3 & 24 million & 48 \\ \hline
    AlexNet & 62 million & 8 \\ \hline
    ResNet-18 & 11 million & 18 \\ \hline
    GoogLeNet & 7 million & 22 \\ \hline
    VoltaVision & 29 thousand & 3 \\ \hline
    \end{tabular}

    \caption{Complexity of the pre-trained models}
    \label{model complexity}
\end{table}

For each of the pre-trained models, the last layer is replaced with a new one which matches the number of desired classes (i.e., 3). Then, the model is fine-tuned for 25 epochs to classify the target components: Bluetooth module, humidity sensor and transistor. The fine-tuning dataset contained around 110 images of each class \ref{sec: dataset}. It is divided into 5 folds to perform 5-fold cross validation where each time 1 fold is selected as the validation set and the rest of the folds form the train set. A step LR scheduler with a step size of 7 and a gamma of 0.1 are used to reduce the learning rate iteratively. As for the optimizer and loss function, SGD and CrossEntropy Loss are used. More complex models like VGG-16 and AlexNet (shown in Table \ref{model complexity}), take much longer time and resources (shown in Table \ref{result_summary}) than VoltaVision to train and thus, produce relatively better results than VoltaVision.
\end{document}